\documentclass[runningheads]{llncs}
\usepackage{graphicx}
\usepackage[]{algorithm2e}
\usepackage{upgreek}
\usepackage{amsmath}
\usepackage{bbold}
\usepackage{cite}
\usepackage{ulem}
\usepackage{graphicx}
\usepackage{grffile}
\graphicspath{ {./Images/} }
\usepackage{multirow}
\usepackage{arydshln}
\usepackage{hhline}
\usepackage{mathtools}

\usepackage{fancyvrb}
\usepackage{color}
\definecolor{cb_orange}{rgb}{1.0,0.51,0.0}
\definecolor{cb_blue}{rgb}{0.22,0.49,0.72}
\definecolor{cb_green}{rgb}{0.3,0.67,0.29}
\definecolor{cb_red}{rgb}{0.89,0.1,0.11}
\definecolor{cb_purple}{rgb}{0.6, 0.31, 0.64}
\definecolor{cadetgrey}{rgb}{0.57, 0.64, 0.69}

\newcommand{\hs}[1]{#1}
\newcommand{\delete}[1]{}

\begin{document}

\title{Scribble2Label: Scribble-Supervised Cell Segmentation via Self-Generating Pseudo-Labels with Consistency} 

%
%

\author{Hyeonsoo Lee\inst{1} \and
Won-Ki Jeong\inst{2}\thanks{Corresponding author: wkjeong@korea.ac.kr}}
\authorrunning{F. Author et al.}
%
\institute{Ulsan National Institute of Science and Technology, \\
School of Electrical and Computer Engineering \\
\email{hslee1@unist.ac.kr} \and
Korea University, College of Informatics, \\
Department of Computer Science and Engineering \\
\email{wkjeong@korea.ac.kr}}

\maketitle              
\begin{abstract}
Segmentation is a fundamental process in microscopic cell image analysis. 
With the advent of recent advances in deep learning, more accurate and high-throughput cell segmentation has become feasible. 
However, most existing deep learning-based cell segmentation algorithms require fully annotated ground-truth cell labels, which are time-consuming and labor-intensive to generate. 
In this paper, we introduce \texttt{Scribble2Label}, a novel weakly-supervised cell segmentation framework that exploits only a handful of scribble annotations without full segmentation labels. 
The core idea is to combine pseudo-labeling and label filtering to generate reliable labels from weak supervision. 
For this, we leverage the consistency of predictions by iteratively averaging the predictions to improve pseudo labels.
%
%
We demonstrate the performance of \texttt{Scribble2Label} by comparing it to several state-of-the-art cell segmentation methods with various cell image modalities, including bright-field, fluorescence, and electron microscopy.
We also show that our method performs robustly across different levels of scribble details, which confirms that only a few scribble annotations are required in real-use cases.


\end{abstract}

\section{Introduction}
\label{introduction}

Micro- to nano-scale microscopy images are commonly used for cellular-level biological image analysis. 
In cell image analysis, segmentation serves as a crucial task to extract the morphology of the cellular structures. 
%
Conventional cell segmentation methods are mostly grounded in model-based and energy minimization methods, such as Watershed~\cite{yang2006nuclei}, Chan-Vese with the edge model~\cite{mavska2013segmentation}, and gradient vector flow~\cite{li2005segmentation}.
%
%
The recent success of deep learning has gained much attention in many image processing and computer vision tasks. 
%
A common approach to achieve highly-accurate segmentation performance is to train deep neural networks using ground-truth labels~\cite{xing2015automatic,arbelle2019microscopy,yi2019multi}. 
However,  generating a sufficient number of ground-truth labels is time-consuming and labor-intensive, which is becoming a major bottleneck in the segmentation process. 
Additionally, manually generated segmentation labels are prone to errors due to the difficulty in drawing pixel-level accurate region masks.

To address such problems, weakly-supervised cell segmentation methods using point annotation have recently been proposed~\cite{nishimura2019weakly,qu2019weakly,yoo2019pseudoedgenet}.
Yoo et al.~\cite{yoo2019pseudoedgenet} and Qu et al.~\cite{qu2019weakly} introduced methods that generate coarse labels only from point annotations using a Voronoi diagram. 
Further, Nishimura et al.~\cite{nishimura2019weakly} proposed a point detection network in which output is used for cell instance segmentation. 
%
Even though point annotation is much easier to generate compared to full region masks, the existing work requires point annotations for the entire dataset -- for example, there are \delete{37,333 nuclei in 841 images of the BBBC0381 dataset~\cite{caicedo2019nucleus} and }around 22,000 nuclei in 30 images of the MoNuSeg dataset~\cite{kumar2017dataset}. 
%
Moreover, the performance of the work mentioned above is highly sensitive to the point location, i.e., the point should be close to the center of the cell. 

Recently, weakly-supervised learning using scribble annotations, i.e., scribble-supervised learning, has actively been studied in image segmentation as a promising direction for lessening the burden of manually generating training labels. 
\delete{The main idea of scribble-supervised learning is to train a segmentation model using a sparse set of scribble labels without a full set of ground-truth masks. 
}
Scribble-supervised learning exploits scribble labels and regularized networks with standard segmentation techniques (e.g., graph-cut~ \cite{lin2016scribblesup}, Dense Conditional Random Field [DenseCRF]~\cite{can2018learning, tang2018regularized}) or additional model parameters (e.g., boundary prediction~\cite{wang2019boundary} and adversarial training~\cite{wu2018scribble}).
The existing scribble-supervised methods have demonstrated the possibility to reduce manual efforts in generating training labels, but their adaptation in cell segmentation has not been explored yet.

In this paper, we propose a novel weakly-supervised cell segmentation method that is highly accurate and robust with only a handful of manual annotations.
Our method, \texttt{Scribble2Label}, uses scribble annotations as in conventional scribble-supervised learning methods, but we propose the combination of pseudo-labeling and label-filtering to progressively generate full training labels from a few scribble annotations.
By doing this, we can effectively remove noise in pseudo labels and improve prediction accuracy.
The main contributions of our work are as follows.
{
\begin{itemize}
	\item We introduce a novel iterative segmentation network training process that generates training labels automatically via weak-supervision using only a small set of manual scribbles, which significantly reduces the manual effort in generating training labels.
	 \item We propose a novel idea of combining pseudo-labeling with label filtering, exploiting consistency to generate reliable training labels, which results in a highly accurate and robust performance. We demonstrate that our method consistently outperforms the existing state-of-the-art methods across various cell image modalities and different levels of scribble details.
	\item Unlike existing scribble-supervised segmentation methods, our method is an end-to-end scheme that does not require any additional model parameters or external segmentation methods (e.g, Graph-cut, DenseCRF) during training.
\end{itemize}
To the best of our knowledge, this is the first scribble-supervised segmentation method applied to the cell segmentation problem in various microscopy images.
}

\section{Method}
\label{method}

\begin{figure}[!t]
\label{fig:label-generation}
\centering
\includegraphics[width=1\textwidth]{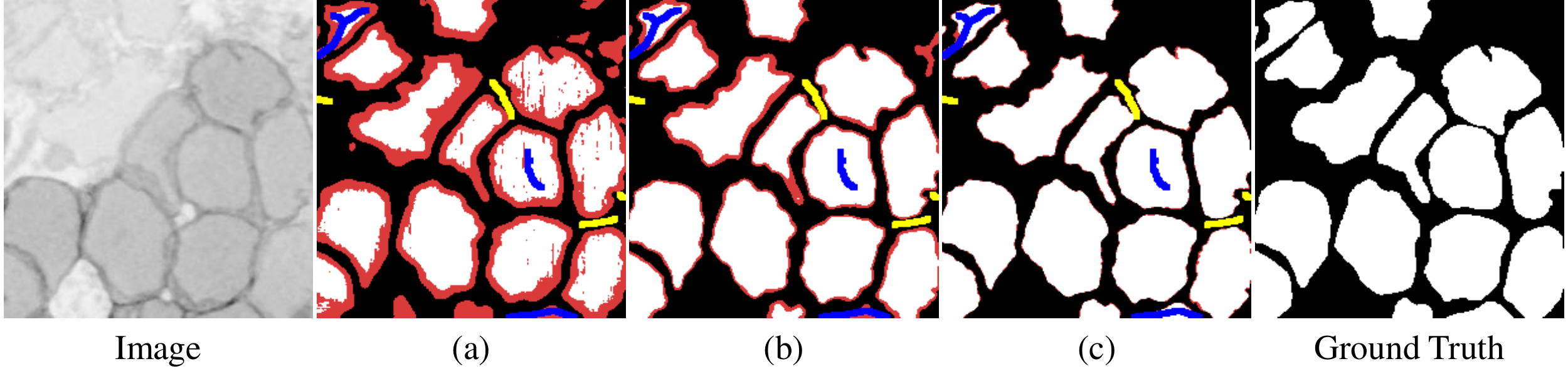}
\caption{An example of iterative refinement of pseudo labels during training. 
Blue and yellow: scribbles for cells and background, respectively ($\Omega_{s}$); red: the pixels below the consistency threshold $\uptau$, which will be ignored when calculating the unscribbled 
pixel loss ($\mathcal{L}_{up}$); white and black: cell or background pixels over $\uptau$ ($\Omega_{g}$). 
\delete{yellow: scribbles for the background; blue: scribbles for cells; and red: the pixels below the consistency threshold $\uptau$, which will be ignored when calculating the unscribbled 
pixel loss ($\mathcal{L}_{up}$). The white and black are the cell or background pixels over $\uptau$ ($\Omega_{g}$). }
%
(a) -- (c) represent the filtered pseudo-labels from the predictions over the iterations (with Intersection over Union [IoU] score): 
(a): 7th (0.5992),
(b): 20th (0.8306), and
(c): 100th (0.9230).
The actual scribble thickness used in our experiment was 1 pixel, but it is widened to 5 pixels in this figure for better visualization. 
}
\end{figure}

\begin{figure}[!t]
\label{fig:overview}
\centering
\includegraphics[width=1\textwidth]{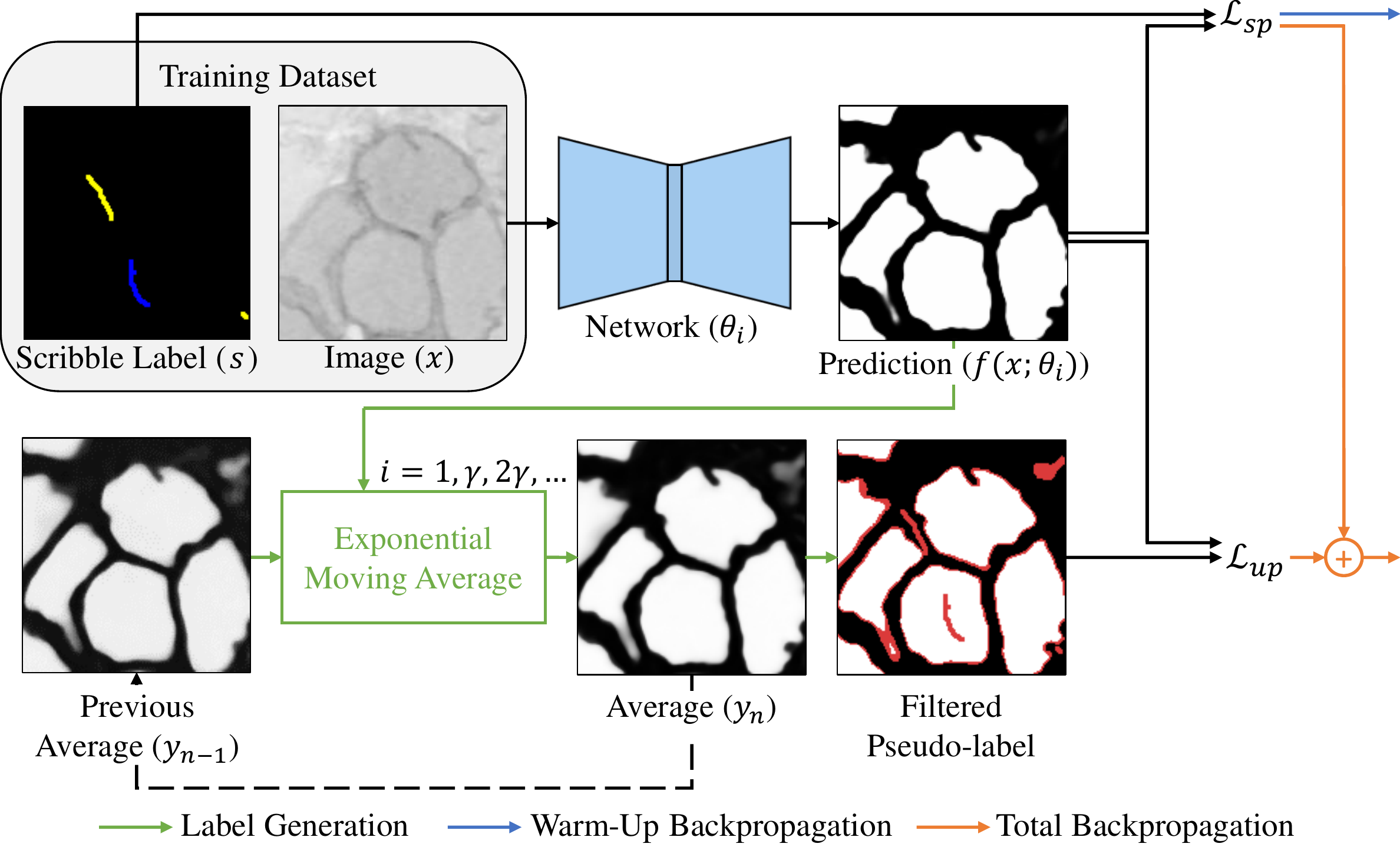}
\caption{The overview of the proposed method (\texttt{Scribble2Label}). The pseudo-label
is generated from the average of predictions. Following, $\mathcal{L}_{sp}$ is calculated with the scribble annotation, and $\mathcal{L}_{up}$ is calculated with the filtered pseudo-label\delete{ (i.e., red pixels are ignored when computing the loss)}. 
%
The prediction ensemble process occurs every $\gamma$ epochs, where $\gamma$ is the ensemble interval.
$n$ represents how many times the predictions are averaged.
}
\label{fig_generation}
\end{figure}

In this section, we describe the proposed segmentation method in detail. 
The input sources for our method are the image $x$ and the user-given scribbles $s$ (see Figure~\ref{fig:overview}). 
Here, the given scribbles are \textit{labeled} pixels (denoted as blue and yellow for the foreground and background, respectively), and the rest of the pixels are \textit{unlabeled} pixels (denoted as black). 
For labeled (scribbled) pixels, a standard cross-entropy loss is applied.
For unlabeled (unscribbled) pixels, our network automatically generates reliable labels using the exponential moving average of the predictions during training. 
Training our model consists of two stages. The first stage is initialization (i.e., a warm-up stage) by training the model using only the scribbled pixel loss ($\mathcal{L}_{sp}$).
Once the model is initially trained via the warm-up stage, the prediction is iteratively refined by both scribbled and unscribbled losses ($\mathcal{L}_{sp}$ and $\mathcal{L}_{up}$). 
Figure~\ref{fig:overview} illustrates the overview of the proposed system. 
%



\subsection{Warm-Up Stage}
\label{method:warm-up}
At the beginning, we only have a small set of user-drawn scribbles for input training data. 
During the first few iterations (warm-up stage), we train the model only using the given scribbles, and generate the average of predictions which can be used in the following stage (Section~\ref{sec:self}).
%
%
%
%
Here, the given scribbles is a subset of the corresponding mask annotation.
By ignoring unscribbled pixels, the proposed network is trained with cross entropy loss as follows:
\begin{equation}
    \mathcal{L}_{sp}(x, s) = - {1 \over |\Omega_{s}|} \sum \limits_{j \in \Omega_{s}} [s_{j}\log(f(x; \theta_{i})) + (1-s_{j})\log(1-f(x; \theta_{i}))] ,
    \label{eq:sp}
\end{equation}
where $x$ is an input image, $s$ is a scribble annotation, and $\Omega_{s}$ is a set of scribbled pixels. 
$f(x; \theta_{i})$ is the model's prediction at iteration $i$.
This warm-up stage continues until we reach the warm-up Epoch $E_{W}$.

Moreover, we periodically calculate the exponential moving average (EMA) of the predictions over the training process: $y_{n} = \alpha f(x;\theta_{i})+(1-\alpha)y_{n-1}$ where $\alpha$ is the EMA weight, 
$y$ is the average of predictions\hs{, $y_{0}=f(x;\theta_{1})$, and $n$ is how many times the predictions are averaged.}
This process is called a prediction ensemble~\cite{nguyen2019self}. 
%
%
Note that, since we use data augmentation for training, the segmentation prediction is not consistent for the same input image. 
Our solution for this problem is splitting the training process into training and ensemble steps.
In the ensemble phase, an un-augmented image is used for the input to the network, and EMA is applied to that predictions.
Moreover, in the scribble-supervised setting, we cannot ensemble the predictions when the best model is found, as in \cite{nguyen2019self}, because the given label is not fully annotated.
To achieve the valuable ensemble and reduce computational costs, the predictions are averaged every $\gamma$ epochs, where $\gamma$ is the ensemble interval.
\delete{Formally, at iteration $i$, $\mathcal{L}_{up}$ is calculated with ($x, y_n$), where $n = \lfloor {i / \gamma} \rfloor + 1$.} 
%

%

\subsection{Learning with a Self-Generated Pseudo-Label}
\label{sec:self}
The average of the predictions can be obtained after the warm-up stage.
\hs{This can be used for generating a reliable pseudo-label of unscirbbled pixels.}
\delete{This can be used as a label for unscirbbled pixels.
However, this average itself is noisy because it comes from the model's prediction.
Thus, we need to filter it.
}
For filtering the pseudo-label, the average is used. The pixels with consistently the same result are
one-hot encoded and used as a label for unscribbled pixels with standard cross entropy.
Using only reliable pixels and making these one-hot encoded progressively provide benefits through curriculum learning and entropy minimization\cite{sohn2020fixmatch}.
With filtered pseudo-label, the unscribbled pixel loss is is defined as follows:
\begin{equation}
    \mathcal{L}_{up}(x, y_{n}) = -{1 \over |\Omega_{g}|} \sum \limits_{j \in \Omega_{g}} [\mathbb{1}(y_{n}>\uptau)\log(f(x;\theta_{i})) + \mathbb{1}((1-y_{n})>\uptau))\log(1-f(x;\theta_{i}))],
    \label{eq:up}   
\end{equation}
where $\Omega_{g} = \{g|g \in (max(y_{n},1-y_{n}) > \uptau), g \not\in \Omega_{s}\}$, which is a set of generated label pixels, and 
$\uptau$ is the consistency threshold. 
Formally, at iteration $i$, $\mathcal{L}_{up}$ is calculated with ($x, y_n$), where $n = \lfloor {i / \gamma} \rfloor + 1$.
The total loss is then defined as the combination of the scribbled loss $\mathcal{L}_{sp}$ and the unscribbled loss $\mathcal{L}_{up}$ with the relative weight of $\mathcal{L}_{up}$, defined as follows:
%
\begin{equation}
    \mathcal{L}_{total}(x, s, y_{n}) = \mathcal{L}_{sp}(x, s) + \lambda \mathcal{L}_{up}(x, y_{n})
    \label{eq:total}
\end{equation}
Note the EMA method shown above is also applied during this training process.

\section{Experiments}
\label{experiments}
%
\subsection{Datasets}

We demonstrated the efficacy of our method using three different cell image datasets.
The first set, MoNuSeg~\cite{kumar2017dataset}, consists of 30 1000$\times$1000 histopathology images acquired from multiple sites covering diverse nuclear appearances. 
We conducted a 10-fold cross-validation for the MoNuSeg dataset. 
BBBC038v1~\cite{caicedo2019nucleus}, the second data set, which is known as Data Science Bowl 2018, is a set of nuclei 2D images.
%
%
We used the stage 1 training dataset, which is fully annotated, and further divided it into three main types, including 542 fluorescence (DSB-Fluo) images of various sizes, 108 320$\times$256 histopathology images (DSB-Histo), and 16 bright-field 1000$\times$1000 (DSB-BF) images.
Each dataset is split into training, validation, and test sets, with ratios of 60\%, 20\%, and 20\%, respectively.
EM is an internally collected serial-section electron microscopy image dataset of a larval zebrafish. 
%
%
\hs{
We used three sub-volumes of either 512$\times$512$\times$512 or 512$\times$512$\times$256 in size. 
The size of the testing volume was 512$\times$512$\times$512.
}
%
The scribbles of MoNuSeg and DSBs were manually drawn by referencing the full segmentation labels. 
To ensure the convenience of scribbles, we annotate images up to 256$\times$256 within 1 min, 512$\times$512 within 2 min, and 1024$\times$1024 within 4 min.
%
%
%
\hs{For the EM dataset, the scribble annotation was generated in the same manner used to evaluate the effect of various amounts of scribble annotations with a 10\% ratio.}

\subsection{Implementation Details}
\begin{figure}[!t]
\centering
\includegraphics[width=1\textwidth]{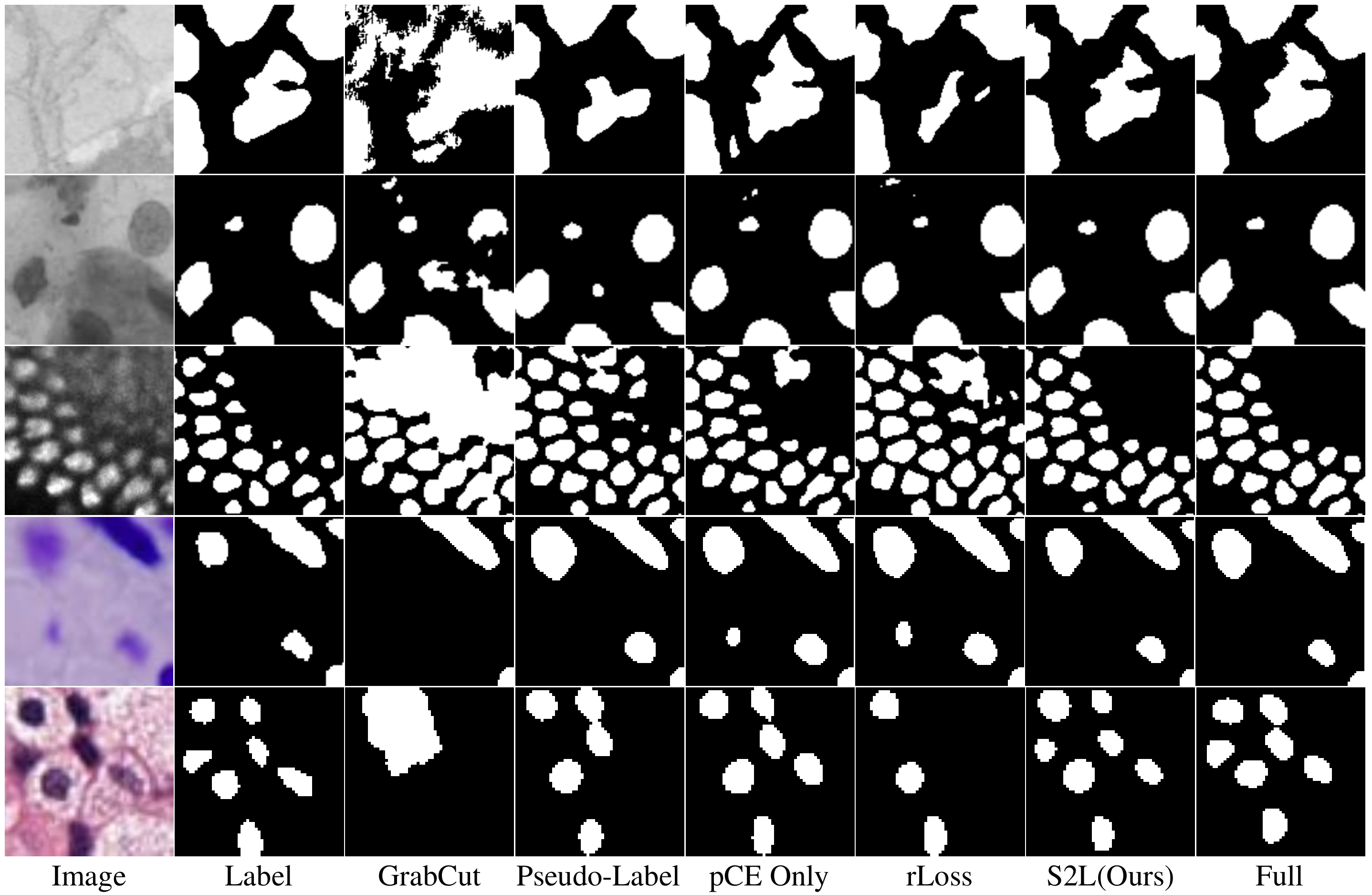}
\caption{Qualitative results comparison. From the top to the bottom, EM,  DSB-BF\cite{caicedo2019nucleus}, DSB-Fluo, DSB-Histo, and MoNuSeg\cite{kumar2017dataset} are shown.}
\label{fig_quality}
\end{figure}
Our baseline network was U-Net~\cite{ronneberger2015u} with the ResNet-50~\cite{he2016deep} encoder. 
For comparison with~\cite{qu2019weakly} in histopathology experiments (MoNuSeg, DSB-Histo), we used ResNet-34 for the encoder. 
The network was initialized with pre-trained parameters, and RAdam~\cite{liu2019variance} was used for all experiments.
To regularize the network, we used conventional data augmentation methods, such as cropping, flipping, rotation, shifting, scaling, brightness change, and contrast changes.

The hyper-parameters used for our model are as follows: Consistency Threshold $\uptau=0.8$; EMA Alpha $\alpha=0.2$; Ensemble Momentum $\gamma=5$; $\mathcal{L}_{up}$'s weight $\lambda=0.5$; and warm-up epoch $E_W=100$.
%
For the MoNuSeg dataset (which is much noisier than other datasets), we used $\uptau=0.95$ and $\alpha=0.1$ to cope with noisy labels.
\subsection{Results}
%
%
We evaluated the performance of semantic segmentation using the intersection over union (IoU) and the performance of instance segmentation using mean Dice-coefficient (mDice) used in \cite{nishimura2019weakly}.
%

\begin{table}[!t]
\caption{Quantitative results of various cell image modalities. The numbers represent accuracy 
in the format of IoU[mDice].}
\centering
\begin{tabular}{|c|c|c|c|c|c|c|} 
\hline
\textbf{Label}       & \textbf{Method} & \textbf{EM}                                                                & \textbf{DSB-BF}                                                            & \textbf{DSB-Fluo}                                                          & \textbf{DSB-Histo}                                                         & \textbf{MoNuSeg}                                                            \\ 
\hline\hline
\multirow{5}{*}{Scribble} & GrabCut\cite{lin2016scribblesup}         & \begin{tabular}[c]{@{}c@{}}0.5288\\{[}0.6066]\end{tabular}                 & \begin{tabular}[c]{@{}c@{}}0.7328\\{[}0.7207]\end{tabular}                 & \begin{tabular}[c]{@{}c@{}}0.8019\\{[}0.7815]\end{tabular}                 & \begin{tabular}[c]{@{}c@{}}0.6969\\{[}0.5961]\end{tabular}                 & \begin{tabular}[c]{@{}c@{}}0.1534\\{[}0.0703]\end{tabular}                  \\ 
\cline{2-7}
                          & Pseudo-Label\cite{lee2013pseudo}    & \begin{tabular}[c]{@{}c@{}}0.9126\\{[}0.9096]\end{tabular}                 & \begin{tabular}[c]{@{}c@{}}0.6177\\{[}0.6826]\end{tabular}                 & \begin{tabular}[c]{@{}c@{}}0.8109\\{[}0.8136]\end{tabular}                 & \begin{tabular}[c]{@{}c@{}}0.7888\\{[}0.7096]\end{tabular}                 & \begin{tabular}[c]{@{}c@{}}0.6113\\{[}0.5607]\end{tabular}                  \\ 
\cline{2-7}
                          & pCE Only~\cite{tang2018regularized}        & \begin{tabular}[c]{@{}c@{}}0.9000\\{[}0.9032]\end{tabular}                 & \begin{tabular}[c]{@{}c@{}}0.7954\\{[}0.7351]\end{tabular}                 & \begin{tabular}[c]{@{}c@{}}0.8293\\{[}0.8375]\end{tabular}                 & \begin{tabular}[c]{@{}c@{}}0.7804\\{[}0.7173]\end{tabular}                 & \begin{tabular}[c]{@{}c@{}}0.6319\\{[}0.5766]\end{tabular}                  \\ 
\cline{2-7}
                          & rLoss\cite{tang2018regularized}           & \begin{tabular}[c]{@{}c@{}}0.9108\\{[}0.9100]\end{tabular}                 & \begin{tabular}[c]{@{}c@{}}0.7993\\{[}0.7280]\end{tabular}                 & \begin{tabular}[c]{@{}c@{}}0.8334\\{[}0.8394]\end{tabular}                 & \begin{tabular}[c]{@{}c@{}}0.7873\\{[}0.7177]\end{tabular}                 & \begin{tabular}[c]{@{}c@{}}0.6337\\{[}0.5789]\end{tabular}                  \\ 
\cline{2-7}
                          & S2L(Ours)       & \begin{tabular}[c]{@{}c@{}}\textbf{0.9208}\\\textbf{[0.9167]}\end{tabular} & \begin{tabular}[c]{@{}c@{}}\textbf{0.8236}\\\textbf{[0.7663]}\end{tabular} & \begin{tabular}[c]{@{}c@{}}\textbf{0.8426}\\\textbf{[0.8443]}\end{tabular} & \begin{tabular}[c]{@{}c@{}}\textbf{0.7970}\\\textbf{[0.7246]}\end{tabular} & \begin{tabular}[c]{@{}c@{}}\textbf{0.6408}\\\textbf{[0.5811]}\end{tabular}  \\ 
\hline\hline
Point                     & \textit{Qu}\cite{qu2019weakly}          & -                                                                          & -                                                                          & -                                                                          & \begin{tabular}[c]{@{}c@{}}0.5544\\{[}0.7204]\end{tabular}                 & \begin{tabular}[c]{@{}c@{}}0.6099\\{[}0.7127]\end{tabular}                  \\ 
\hline\hline
Full                      & Full            & \begin{tabular}[c]{@{}c@{}}0.9298\\{[}0.9149]\end{tabular}                 & \begin{tabular}[c]{@{}c@{}}0.8774\\{[}0.7879]\end{tabular}                 & \begin{tabular}[c]{@{}c@{}}0.8688\\{[}0.8390]\end{tabular}                 & \begin{tabular}[c]{@{}c@{}}0.8134\\{[}0.7014]\end{tabular}                 & \begin{tabular}[c]{@{}c@{}}0.7014\\{[}0.6677]\end{tabular}                  \\
\hline
\end{tabular}
\label{tab:dataset}
\end{table}
\label{table_dataset}
\textbf{Comparison with other methods: }
%
We compared our method to the network trained with full segmentation annotation, scribble annotation (pCE Only)~\cite{tang2018regularized}, and the segmentation proposal from Grab-Cut~\cite{lin2016scribblesup}.
%
%
To demonstrate the efficacy of the label filtering with consistency, 
we compared it to pseudo-labeling~\cite{lee2013pseudo}. 
The pixels for which the probability of prediction were over threshold $\uptau$ were assigned to be a pseudo-label, where $\uptau$ was same as our method setting.
%
%
Our method was also compared to Regularized Loss (rLoss)\cite{tang2018regularized}, which integrates the DenseCRF into the loss function. The hyper-parameters of rLoss are $\sigma_{XY}=100$ and $\sigma_{RGB}=15$.

Table \ref{tab:dataset} shows the quantitative comparison of our method with several representative \delete{cell segmentation }methods. 
Overall, our method outperformed all methods on both IoU and mDice quality metrics. 
We observed that our method achieved even higher mDice accuracy compared to the full method (i.e., trained using full segmentation labels) on EM, DSB-BF, and DSB-Histo datasets.
Note also that MoNuSeg dataset contains many small cluttering cells, which are challenge to separate individually. 
However, our method showed outstanding instance segmentation results in this case, too.


%
Grab-Cut's\cite{lin2016scribblesup} segmentation proposal \hs{and the pseudo-label~\cite{lee2013pseudo}} were erroneous. Thus, training with these erroneous segmentation labels impairs 
the performance of the method.
%
%
Qu et al.'s method~\cite{qu2019weakly} performed well for instance-level segmentation on MoNuSeg dataset, however, it performed worse on DSB-histo dataset. 
Because \cite{qu2019weakly} used a clustering label that has circular shape cell label, it was hard to segment the non-circular cell.
Learning with pCE~\cite{tang2018regularized} showed stable results on various datasets. 
However, due to learning using only scribbles, the method failed to correctly predict boundary accurately as in our method. 
%
%
rLoss~\cite{tang2018regularized} outperformed most of the previous methods, but our method 
generally showed better results.
We also observed that leveraging consistency by averaging predictions is crucial to generate robust pseudo-labels.
%
{Scribble2Label}'s results also confirm that using pseudo label together with scribbles is effective to generate accurate boundaries, comparable to the ground-truth segmentation label.



%
\begin{table}[!t]
\caption{Quantitative results using various amounts of scribbles. DSB-Fluo\cite{caicedo2019nucleus} was used for the evaluation. The numbers represent accuracy in the format of IoU[mDice].}
\centering
\begin{tabular}{|c|c|c|c|c||c|} 
\hline
\textbf{Method} & \textbf{10\%}                                                              & \textbf{30\%}                                                              & \textbf{50\%}                                                              & \textbf{100\%}                                                             & \textbf{Manual}                                                            \\ 
\hline\hline
GrabCut\cite{lin2016scribblesup}         & \begin{tabular}[c]{@{}c@{}}0.7131\\{[}0.7274]\end{tabular}                 & \begin{tabular}[c]{@{}c@{}}0.8153\\{[}0.7917]\end{tabular}                 & \begin{tabular}[c]{@{}c@{}}0.8244\\{[}0.8005]\end{tabular}                 & \begin{tabular}[c]{@{}c@{}}0.8331\\{[}0.8163]\end{tabular}                 & \begin{tabular}[c]{@{}c@{}}0.8019\\{[}0.7815]\end{tabular}                 \\ 
\hline
Pseudo-Label\cite{lee2013pseudo}    & \begin{tabular}[c]{@{}c@{}}0.7920\\{[}0.8086]\end{tabular}                 & \begin{tabular}[c]{@{}c@{}}0.7984\\{[}0.8236]\end{tabular}                 & \begin{tabular}[c]{@{}c@{}}0.8316\\{[}0.8392]\end{tabular}                 & \begin{tabular}[c]{@{}c@{}}0.8283\\{[}0.8251]\end{tabular}                 & \begin{tabular}[c]{@{}c@{}}0.8109\\{[}0.8136]\end{tabular}                 \\ 
\hline
pCE Only~\cite{tang2018regularized}        & \begin{tabular}[c]{@{}c@{}}0.7996\\{[}0.8136]\end{tabular}                 & \begin{tabular}[c]{@{}c@{}}0.8180\\{[}0.8251]\end{tabular}                 & \begin{tabular}[c]{@{}c@{}}0.8189\\{[}0.8204]\end{tabular}                 & \begin{tabular}[c]{@{}c@{}}0.8098\\{[}0.8263]\end{tabular}                 & \begin{tabular}[c]{@{}c@{}}0.8293\\{[}0.8375]\end{tabular}                 \\ 
\hline
rLoss\cite{tang2018regularized}           & \begin{tabular}[c]{@{}c@{}}0.8159\\{[}0.8181]\end{tabular}                 & \begin{tabular}[c]{@{}c@{}}0.8251\\{[}0.8216]\end{tabular}                 & \begin{tabular}[c]{@{}c@{}}0.8327\\{[}0.8260]\end{tabular}                 & \begin{tabular}[c]{@{}c@{}}0.8318\\{[}0.8369]\end{tabular}                 & \begin{tabular}[c]{@{}c@{}}0.8334\\{[}0.8394]\end{tabular}                 \\ 
\hline
S2L(Ours)       & \begin{tabular}[c]{@{}c@{}}\textbf{0.8274}\\\textbf{[0.8188]}\end{tabular} & \begin{tabular}[c]{@{}c@{}}\textbf{0.8539}\\\textbf{[0.8407]}\end{tabular} & \begin{tabular}[c]{@{}c@{}}\textbf{0.8497}\\\textbf{[0.8406]}\end{tabular} & \begin{tabular}[c]{@{}c@{}}\textbf{0.8588}\\\textbf{[0.8443]}\end{tabular} & \begin{tabular}[c]{@{}c@{}}\textbf{0.8426}\\\textbf{[0.8443]}\end{tabular}  \\ 
\hline\hline
Full            & \multicolumn{5}{c|}{\begin{tabular}[c]{@{}c@{}}0.8688\\{[}0.8390]\\\end{tabular}}                                                                                                                                                                                                                                                                                                              \\
\hline
\end{tabular}
\label{tab:scribble}
\end{table}
\textbf{Effect of amount of scribble annotations:}
To demonstrate the robustness of our method over various levels of scribble details, 
we conducted an experiment using 
scribbles automatically generated using a similar method by Wu et al.~\cite{wu2018scribble} (i.e., foreground and background regions are skeletonized and sampled). 
%
%
The target dataset was DSB-Fluo, and various amounts of scribbles, i.e., 10\%, 30\%, 50\%, and 100\% of the skeleton pixels extracted from the full segmentation labels (masks), are automatically generated. 
%
%
%
%
%
Table \ref{tab:scribble} summarizes the results with different levels of scribble details. 
Our method \texttt{Scribble2Label} generated stable results in both the semantic metric and instance metric from sparse scribbles to abundant scribbles.
%

%

The segmentation proposal from Grab-Cut~\cite{lin2016scribblesup} and the pseudo-lable~\cite{lee2013pseudo} were noisy in settings lacking annotations, which resulted in degrading the performance.
%
rLoss~\cite{tang2018regularized} performed better than the other methods, but it sometimes failed to generate correct segmentation results especially when the background is complex (causing confusion with cells). 
%
%
%
Our method showed very robust results over various scribble amounts.
Note that our method performs comparable to using full segmentation masks only with 30\% of skeleton pixels.

\section{Conclusion}
\label{conclusion}

In this paper, we introduced~\texttt{Scribble2Label}, a simple but effective scribble-supervised learning method that combines pseudo-labeling and label-filtering with consistency. 
Unlike the existing methods, \texttt{Scribble2Label} demonstrates highly-accurate segmentation performance on various datasets and at different levels of scribble detail without extra segmentation processes or additional model parameters.
We envision that our method can effectively avoid time-consuming and labor-intensive manual label generation, which is a major bottleneck in image segmentation.
In the future, we plan to extend our method in more general problem settings other than cell segmentation, including semantic and instance segmentation in images and videos. 
Developing automatic label generation for the segmentation of more complicated biological features, such as tumor regions in histopathology images and mitochondria in nano-scale cell images, is another interesting future research direction. 
\sloppy
\textbf{Acknowledgements.} This work was partially supported by the Bio \& Medical Technology Development Program of the National Research Foundation of Korea (NRF) funded by the Ministry of Science and ICT (NRF-2015M3A9A7029725, NRF-2019M3E5D2A01063819), and the Korea Health Technology R\&D Project through the Korea Health Industry Development Institute (KHIDI), funded by the Ministry of Health \& Welfare, Republic of Korea (HI18C0316).



%
%
%
\bibliographystyle{splncs04}
\bibliography{refs}

\end{document}